\documentclass[11pt, a4paper, logo, twocolumn, copyright, nonumbering]{googledeepmind}

\usepackage[authoryear, sort&compress, round]{natbib}
\usepackage{listings}
\usepackage{color}
\usepackage{colortbl}
\usepackage{tabularx}
\usepackage{subcaption}
\definecolor{llmcolor}{RGB}{230,230,250}

\title{Simulation Streams: A Programming Paradigm for Controlling Large Language Models and Building Complex Systems with Generative AI.}

\correspondingauthor{sunehag@google.com}

\paperurl{}

\reportnumber{}

\renewcommand{\today}{2025-01-29}

\author[1]{Peter Sunehag}
\author[1]{Joel Z. Leibo}

\affil[1]{Google DeepMind}

\begin{abstract}
We introduce Simulation Streams, a programming paradigm designed to efficiently control and leverage Large Language Models (LLMs) for complex, dynamic simulations and agentic workflows. Our primary goal is to create a \textbf{minimally interfering framework} that harnesses the agentic abilities of LLMs while addressing their limitations in maintaining consistency, selectively ignoring/including information, and enforcing strict world rules. Simulation Streams achieves this through a state-based approach where variables are modified in sequential steps by ``operators,'' producing output on a recurring format and adhering to consistent rules for state variables. This approach focus the LLMs on defined tasks, while aiming to have the context stream remain ``in-distribution''. The approach incorporates an Entity-Component-System (ECS) architecture to write programs in a more intuitive manner, facilitating reuse of workflows 
across different components and entities. This ECS approach enhances the modularity of the output stream, allowing for complex, multi-entity simulations while maintaining format consistency, information control, and rule enforcement. It is supported  by a custom editor that aids in creating, running, and analyzing 
simulations. We demonstrate the versatility of simulation streams through an illustrative example of an ongoing market economy simulation, a social simulation of three characters playing a game of catch in a park and a suite of classical reinforcement learning benchmark tasks. These examples showcase Simulation 
Streams' ability to handle complex, evolving scenarios over 100s-1000s of iterations, facilitate 
comparisons between different agent workflows and models, and maintain consistency and continued interesting developments in LLM-driven simulations.
\end{abstract}

\begin{document}

\maketitle

\section{Introduction}

Large Language Models (LLMs) have demonstrated impressive capabilities in generating complex simulations \citep{park2023generative, wu2023autogen, chen2023agentverse}. Most straightforwardly, simulations can be produced just by prompting an LLM chatbot with instructions to simulate, and to do so using a specific format as this helps specify what is expected from the simulation and if more detailed even "programs" it. The format bears upon what the explicit state of the simulation contains, e.g. time and the location and activities of a number of characters, and the format can help track the state as well as guide attention to preceding entries.
In this way, one can obtain sophisticated, emergent narratives which benefit from the LLM's vast knowledge base. Further, LLM simulations can enable new possibilities - for instance, in a situation like in Fig.~\ref{fig:example-output-stream}, humans could inquire about an agent's progress or agents could request human assistance. 

However, this approach to LLM-based simulations faces the following limitations:
\begin{enumerate}
    \item Maintaining a consistent format over extended iterations cannot be guaranteed and the chatbot can become "lazy" and sum up instead, e.g. by "and so they continued for iterations 5-10 and then everyone went home and would always remember that day".
    \item It's difficult to selectively hide/include information from the LLM, since they are sampled autoregressively and all previous context is typically available for each generation step.
    \item Enforcing strict world rules is challenging, as the LLM may violate established constraints or introduce inconsistencies, e.g. an agent hallucinating that it achieved its objective.
    \item The LLM can fail to properly track the state of the simulation, e.g. what is in an inventory.
    \item For a very long simulation, all the above can lead to compounding inconsistencies and the simulation going far off track.
\end{enumerate}

While alternative approaches to simulation exist, they face their own significant challenges. Traditional agent-based platforms like NetLogo \citep{wilensky1999netlogo} and artificial life simulators like Avida \citep{ofria2004avida} require explicit rule specification through cellular automata or genetic algorithms, making them brittle when faced with novel scenarios. Physics-based simulation engines like MuJoCo \citep{todorov2012mujoco} excel at physical dynamics but struggle with high-level decision making. Reinforcement Learning (RL) \citep{brockman2016openai} including Multi-Agent RL \cite{du2023review}. can learn policies through environment interaction, but require substantial training data and often fail to capture the nuanced reasoning that LLMs handle naturally. FLAME \citep{chopra2023flame} implements differentiable agent-based models with GPU acceleration, enabling gradient-based parameter calibration of parameterized rules.

Recent work has explored various approaches to multi-agent LLM simulations. \citet{park2023generative} introduced generative agents that maintain memory and plan actions, while \citet{wu2023autogen} developed structured conversational flows between specialized agents. \citet{chen2023agentverse} focused on role-playing scenarios with defined social protocols, and \citet{qian2024chatdev} demonstrated specialized agents collaborating in software development. Each of these frameworks implements specific patterns for managing agent interactions.

This paper introduces Simulation Streams\footnote{Implementation available at:\\ \url{github.com/google-deepmind/simulation_streams}}, a programming paradigm designed to leverage the agentic abilities of LLMs while addressing their core limitations in maintaining consistency and enforcing constraints. Importantly, Simulation Streams provides a foundational language that can express many mechanisms of existing multi-agent LLM frameworks while requiring minimal interference with the model's capabilities.

Simulation Streams takes a state-based approach where variables are modified in sequential steps, each defined by what we term an ``operator,'' producing a structured output stream that adheres to consistent rules. The following example in Figure \ref{fig:example-output-stream} demonstrates how Simulation Streams generates a consistent output format across multiple iterations, with different operators contributing to various parts of the stream:

\begin{figure}[h!]
\caption{Example output stream (trajectory) with LLM-generated rows \colorbox{llmcolor}{highlighted in this color}. See Table 1 for the operator definitions used for the simulation from which these timesteps were sampled.}
\label{fig:example-output-stream}
\begin{center}
\begin{tabular}{l}
time = 1 \\
objective = "Find cheese in 5x5 grid" \\
high\_level\_plan = "Explore systematically" \\
movement\_plan = "Move right, then down" \\
move\_x = 1 \\
move\_y = 0 \\
location\_x = 1 \\
location\_y = 0 \\
previously\_searched = [(0, 0), (1, 0)] \\
cheese\_found = False \\
summary = "At (1, 0). No cheese.  \\
Searched: [(0, 0), (1, 0)].  \\
Just started, cheese could be anywhere." \\
\\
time = 2 \\
\rowcolor{llmcolor} high\_level\_plan = "Continue exploration" \\
\rowcolor{llmcolor} movement\_plan = "Move downwards" \\
\rowcolor{llmcolor} move\_x = 0 \\
\rowcolor{llmcolor} move\_y = 1 \\
location\_x = 1 \\
location\_y = 1 \\
previously\_searched = [(0, 0), (1, 0), (1, 1)] \\
cheese\_found = False \\
\rowcolor{llmcolor} summary = "At (1, 1). No cheese yet.  \\
\rowcolor{llmcolor} Searched: [(0, 0), (1, 0), (1, 1)].  \\
\rowcolor{llmcolor} Covered 3 of 25 cells, still early exploration." \\
\end{tabular}
\end{center}
\end{figure}

 A crucial aspect of leveraging LLMs in Simulation Streams is maintaining "in-distribution" generation, which largely means generating a context where everything is reasonably probably according to the LLM. We want the trajectory to be something it could have generated by itself, while we are forcing it to be within the subset of trajectories it could have generated that follows our rules. In the context of LLMs, remaining in-distribution refers to generating content that aligns closely with the patterns and structures the model has learnt in training. To achieve this, Simulation Streams organizes its output into substreams---sequences of related entries that follow consistent, well-defined formats. Each substream maintains its own specific structure, providing a clear pattern of manageable complexity for the LLM to work with, like the example in Figure \ref{fig:example-output-stream}. When generating new content, the LLM is tasked with producing the next line in the relevant substream, following the established format.

One of the key advantages of the substream approach is its ability to keep the repeating format and its events sufficiently simple and compact. This simplicity allows the LLM to model the pattern more effectively, enabling it to attend more appropriately to relevant preceding entries. By breaking the simulation into substreams, we create a modular design that is easier to manage and modify, while also improving the LLM's ability to generate coherent and contextually relevant content. As simulations grow in complexity, the substream approach allows for the addition of new elements without disrupting existing patterns, enhancing scalability.

Programs in Simulation Streams are essentially lists of operators, defining formats and rules while allowing easy switching between human design and LLM generation based on provided conditions. The simulation maintains a state composed of variables that can be modified at each step and influence the output. Each step in the simulation is defined by an operator, which includes an unique identifier, a formula (expressed in Python) that describes how variables change, conditions for when the LLM should generate or modify the formula, a query that determines which substreams form the context for the operator, and a "next" field that determines the subsequent operator. Many of these fields are provided automatically most of the time.

Formulas within operators can be either human-written or generated by LLMs, allowing for both precise control and dynamic adaptability in the simulation. This flexibility is crucial for creating complex, evolving scenarios that can adapt to unexpected situations or novel inputs. The simulation progresses by evaluating operators sequentially, with each step potentially influencing which operator is executed next. Each step appears as a row in the stream, showing the variable being assigned and the right-hand side that determines its new value. This provides a traceable record of the LLM's behavior and outputs, facilitating debugging and analysis of the simulation's progress. Each operator's execution results in a contribution to the output stream, either through direct state updates or LLM-generated content. This structured approach allows Simulation Streams to maintain format consistency over thousands of iterations if needed, selectively control the LLM's access to information, and strictly enforce simulation rules - addressing the key limitations of purely LLM-driven simulations.

To further enhance the usability, Simulation Streams incorporates an Entity-Component-System (ECS) architecture \citep{bilas2002,fabian2009}
 for producing lists of operators, which form the programs. This approach allows for modular, reusable components in building complex LLM-driven simulations. The ECS architecture is supported by a custom editor that has been developed to facilitate the creation of operator lists, run simulations, inspect metrics, and allow querying of the generated simulation stream. This ECS approach enhances the modularity of the output stream, allowing for complex, multi-entity simulations while maintaining the core benefits of format consistency, information control, and rule enforcement.

To demonstrate the versatility and power of Simulation Streams, we provide an illustrative example of simulating a market economy with consumers, producers of various goods, a market maker, and common resources. This simulation runs indefinitely, generating ongoing interesting and relevant developments. Additionally, we create a suite of classical reinforcement learning benchmark tasks, including grid world tasks and continuous control problems like mountain car, where agent workflows and different models can be compared.

\newcommand{\stackedtags}[1]{\begin{tabular}[t]{@{}c@{}}#1\end{tabular}}

\begin{table*}[ht]
\footnotesize
\begin{tabularx}{\textwidth}{|l|X|c|c|c|}
\hline
\textbf{Operator} & \textbf{Formula/Action} & \textbf{LLM Condition} & \textbf{Query} & \textbf{Tags} \\
\hline
Time & $time = time + 1$ & False & N/A & All \\
\hline
Objective & $objective = \text{"Find cheese in 5x5 grid"}$ & False & N/A & All \\
\hline
High-Level Plan & $high\_level\_plan = \text{"Explore the grid systematically"}$ & $time > 1$ & $planning = True$ & planning \\
\hline
Movement Plan & $movement\_plan = \text{"Move right then down, repeating"}$ & $time > 1$ & $planning = True$ & \stackedtags{planning \\ movement} \\
\hline
Move\_X & $move\_x = 1 \text{ if } (time \% 2 == 1) \text{ else } 0$ & $time > 1$ & $movement = True$ & movement \\
\hline
Move\_Y & $move\_y = -1 \text{ if } (time \% 2 == 0) \text{ else } 0$ & $time > 1$ & $movement = True$ & movement \\
\hline
Location\_X & $location\_x = \max(0, \min(4, location\_x + move\_x))$ & False & N/A & summary \\
\hline
Location\_Y & $location\_y = \max(0, \min(4, location\_y + move\_y))$ & False & N/A & summary \\
\hline
Previous & $previous.append((location\_x, location\_y))$ & False & N/A & summary \\
\hline
Cheese\_Found & $cheese\_found = cheese\_map[location\_x][location\_y]$ & False & N/A & summary \\
\hline
Summary & $summary = first\_summary$ & $time > 1$ & $summary = True$ & \stackedtags{planning \\ summary} \\
\hline
\end{tabularx}
\caption{Operators for a cheese-finding simulation that generates output as shown in Figure \ref{fig:example-output-stream}. The "How to write a program" section demonstrates the initialization of state variables (e.g., $move\_x$ and $time$), implementation of operators, and their organization into entities (world, agent) and components (planning, movement, summary) - a structure that becomes crucial for complex simulations like the market economy.}
\end{table*}

\section{Simulation Streams}

Simulation Streams is a novel programming paradigm designed for constructing and 
executing complex, dynamic simulations with integrated Large Language Model (LLM) 
capabilities. Formally, we define a Simulation Stream as follows:

A Simulation Stream $S$ is a tuple $(X, O, L, T)$ where:
\begin{itemize}
    \item $X$ is the state space,
    \item $O$ is a set of operators,
    \item $L$ is an LLM sampling function
    \item $T$ is a set of termination conditions.
\end{itemize}
Additionally, an output stream $R$ is maintained, consisting of text rows representing assignment formulas derived from operator executions. Key characteristics that distinguish Simulation Streams include:
\begin{enumerate}
    \item State-based representation of the simulation environment
    \item Operator-driven state transitions
    \item LLM integration
    \item Output stream generation for context
    \item Substream-based context management
\end{enumerate}

\subsection{State Space}

The state space $X$ in a Simulation Stream represents all possible configurations 
of the simulation at any given time. We define it as follows:

The state space $X$ is a set of all possible states $x$, where each state 
$x = (v_1, v_2, ..., v_n)$ is a tuple of state variables $v_i \in V_i$, and 
$V_i$ is the domain of the $i$-th state variable.
State variables can be of various types:
\begin{itemize}
    \item Numerical: $v_i \in \mathbb{R}$ or $v_i \in \mathbb{Z}$ (e.g., real-valued coordinates or integer counters)
    \item Categorical: $v_i \in \{c_1, c_2, ..., c_k\}$ (e.g., boolean values where $v_i \in \{\text{True}, \text{False}\}$, or multi-class labels)
    \item Text: $v_i \in \Sigma^*$, where $\Sigma$ is a finite alphabet (e.g. ASCII characters)
\end{itemize}

The state transition function $F: X \times O \times L \rightarrow X$ defines how 
the state evolves over time:

\begin{equation}
x_{t+1} = F(x_t, o_t, l_t)
\end{equation}

where $x_t$ is the state at time $t$, $o_t \in O$ is the operator applied at time $t$, 
and $l_t$ is the LLM-generated content at time $t$. The selection of the next operator to apply is determined by the next operator selection function $n: O \rightarrow O$, which is a component of each operator. 

\begin{equation}
o_{t+1} = n(o_t).
\end{equation}

Each state transition produces an output in the form of a text row in the output stream $R$. This row comprises an assignment formula where the left-hand side (LHS) coincides with the variable being modified, and the right-hand side (RHS) is the result of evaluating the formula field in the corresponding operator.

\subsection{Operators}

Operators are the mechanism for state transitions in Simulation Streams. 
An operator is defined as follows:

An operator $o \in O$ is defined by the following components:
\begin{itemize}
    \item $id$: a unique identifier,
    \item $f: X \rightarrow X$: a state transition formula that assigns a new value to one variable,
    \item $c: X \rightarrow \{0, 1\}$: an LLM invocation condition,
    \item $q: X \rightarrow \mathcal{P}(X)$: a context query function,
    \item $n: X \rightarrow O$: a next operator selection function,
    \item $A: X \rightarrow X$: a set of additional variable assignments, which can include definitions for sub-stream belongings (tags).
\end{itemize}

The execution of an operator not only transitions the state but also generates a row in the output stream, reflecting the state change performed by the formula in a textual format as described above.

\begin{equation}
f(x) = (v_1, v_2, ..., v_i + \Delta, ..., v_n)
\end{equation}

The LLM invocation condition $c$ determines when the LLM should be used to generate 
or modify the formula. It's a boolean function of the current state:

\begin{equation}
c(x) = \begin{cases} 
\text{True} & \text{if LLM should be invoked} \\
\text{False} & \text{otherwise}
\end{cases}
\end{equation}

\subsection{Output Stream and Query Function}

The output stream $O = (o_1, o_2, ..., o_t)$ is a sequence where each $o_i$ 
represents the output at time step $i$, typically in the form of a text string, e.g.
\begin{verbatim}
'move_x = 1\n'
\end{verbatim}

The context query function $q$ operates on this output stream, selecting 
relevant rows based on specified conditions that relate to the state variables. 
It is defined as follows:

A query function $q: C \times S \times O \rightarrow \mathcal{P}(O)$ takes a 
condition $c \in C$, the current state $s \in S$, and the current output 
stream $O$, and returns a subset of output rows where the condition is true 
based on the state:

\begin{equation}
q(c, s, O) = \{o \in O : c(s, o) \text{ is true}\}
\end{equation}

where $C$ is the set of all possible conditions and $S$ is the state space.

The query function operates on the output stream, which contains the textual representation of state changes. This allows for context selection based on both the current state and the history of state transitions as recorded in the output stream. For example, a query might select all output rows where a movement was generated. 

\begin{equation}
\begin{split}
q(\text{movement=True}, s, O) = \{o_t \in O : \\
    s_t(\text{movement}) = \text{True}\}
\end{split}
\end{equation}

This allows the system to provide relevant historical context for LLM queries 
by selecting specific outputs from the simulation history. In practice, this might look like the following output from executing four operators (and a 'blank' for empty line) where movement=True, in two iterations and we are just about to sample the last of those (for y-movement):

\begin{verbatim}
time = 0
plan = "Move right and down."
move_x = 1
move_y = -1

time = 1
plan = "Move down"
move_x = 0
\end{verbatim}

Here, 'movement' is a state variable that has been set as an additional field 
in some operators and otherwise to a default (False) when not explicitly specified in the operator.

\subsection{Entity-Component-System (ECS)}

Simulation Streams can be organized using an Entity-Component-System (ECS) 
architecture, which provides a modular and hierarchical structure for complex 
simulations. In the context of Simulation Streams, we define ECS as follows:

An ECS architecture in Simulation Streams is a pair $(E, C)$ where:
\begin{itemize}
    \item $E$ is an ordered set of entities, each associated with an ordered list of component identifiers,
    \item $C$ is a set of components, where each component initializes state variables and provides a list of operators.
\end{itemize}

Entities represent distinct objects or actors in the simulation. Formally:

\begin{equation}
E = (e_1, e_2, ..., e_n)
\end{equation}

where each entity $e_i$ is associated with an ordered list of components:

\begin{equation}
e_i \rightarrow [c_{i1}, c_{i2}, ..., c_{ik_i}] \text{ where } c_{ij} \in C
\end{equation}

Components are defined as tuples that include both initial state variables and associated operators:

\begin{equation}
C = \{c_1, c_2, ..., c_m\} \text{ where } c_i = (V_i, O_i)
\end{equation}

Here, $V_i$ is a set of initial values of some state variables, and $O_i$ is a list of operators associated with the component.

The state initialization for an entity $e_j$ is defined by its ordered components:

\begin{equation}
x_0(e_j) = \bigcup_{i=1}^{k_j} V_i \text{ where } c_i \text{ is the } i\text{-th component of } e_j
\end{equation}

The complete list of operators for the simulation is formed by concatenating 
the operators from all components of all entities, respecting the order of 
entities and the order of components within each entity:

\begin{equation}
O = [O_{11}, O_{12}, ..., O_{1k_1}, O_{21}, ..., O_{nk_n}]
\end{equation}

where $O_{ij}$ is the list of operators from the $j$-th component of the $i$-th entity.

The state transition function remains similar to the original definition, but 
now operates on this hierarchically defined set of operators:

\begin{equation}
x_{t+1} = F(x_t, o_t, l_t)
\end{equation}

where $o_t$ is selected from the ordered list $O$.

This ECS structure in Simulation Streams allows for a modular and hierarchical organization of state variables and operators. It provides a flexible framework for defining complex, multi-entity simulations while maintaining the core principles of Simulation Streams. The ordered nature of components within entities ensures a well-defined execution flow.

\section{Experiments}
The purpose of introducing Simulation Streams is to introduce a direct and effective of running long streams of consistent LLM streams that takes the general form of simulations, even if nothing prevent an operator from being based on interaction with an external environment. This experimental section is showing a few examples including for what simulations could be, and we display how it can be used to compare different models. The most complex example is our long-running market economy simulation where you provide an example of how it can run indefinitely and how novel situations and behaviours keep appearing while adhering to reasonable market behaviours, most clearly seen in the economic cycles that is a ubiquitous pattern. 

A conclusion regarding the performance of the models, is that the recent Gemini-2.0-Flash-Exp model and its thinking version, are eminently suited for the simulation streams framework as it is consistently understanding exactly what they need to produce next. This makes the simulations fast, economical and easy to work with, e.g. Gemini-2.0-Flash shows perfect simulation consistency for the social simulation across 10 runs, even if the much larger Gemini-1.5-pro-002 slightly outperforms it on the RL suite. Simulation Streams is based on the belief that improving models is increasingly capable of running agentic simulations without interference, and then benefits from being used in a minimally interfering manner.

\subsection{Results and Analysis}

\subsection{RL tasks}
We used our approach to compare different models on six benchmark tasks inspired by the classical reinforcement learning literature \citep{sutton2018reinforcement}. The tasks include: (1) Windy Gridworld, where an agent must navigate towards a goal while affected by directional wind forces, (2) Key-Chest, where the agent must find and collect a key before accessing a chest, (3) Maze navigation, (4) Mountain Car, where an under-powered car must escape a valley by building momentum, (5) Temperature Control involving maintaining a target temperature within constraints, and (6) Robot Cleaning, where a robot must efficiently clean dirty spots while avoiding obstacles. 

The format employed includes multiple fields: a summary, a high-level strategic plan subject to selective revision, a detailed implementation plan, action determination and consequent state updates. The initial block before LLM invocation serves as an exemplar, establishing foundational patterns that guide subsequent iterations. These initial examples and their default structures significantly impact the system's trajectory.
Maintaining consistency across the initial example fields is essential, and the initial iteration demonstrates how to refer to previously revealed information. The high-level plan functions as the agent's primary agenda, and the initial plan can incorporate various task-relevant strategies. This is exemplified in the mountain car scenario, where the initial plan may include the recognition that alternating backward and forward movements are necessary to generate sufficient momentum before reaching the uphill goal region.

\begin{figure}[t]
   \centering
   \includegraphics[width=\columnwidth]{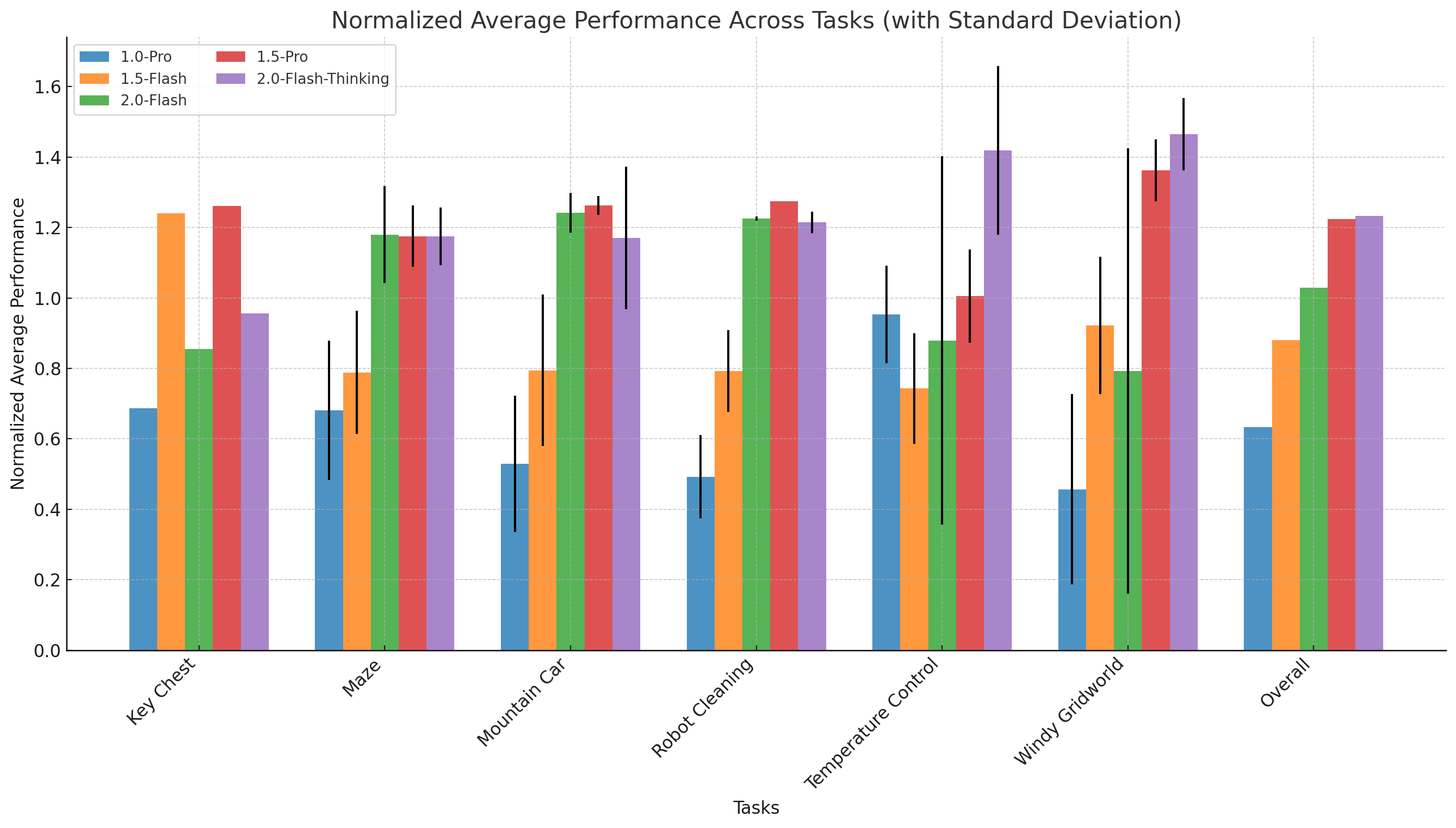}
   \caption{Comparison of performance across six classical reinforcement learning benchmark tasks. Solid lines represent mean performance over 10 runs, with shaded regions indicating standard deviation.}
   \label{fig:comparison}
\end{figure}

We conducted $10$ independent runs of $25$ steps each across the $6$ benchmark tasks. For each task, we collect performance metrics over these runs. The results are presented in Figure~\ref{fig:comparison}, where we show both the mean performance (solid lines) and the standard deviation (shaded regions) across all $10$ runs. We leave out the standard deviation for the Key Chest tasks as performance variability is here highly related to varying key positions. 

Among the compared models, Gemini-1.5-Pro-002 consistently outperforms Gemini-1.5-Flash-002 across all six tasks, while Gemini-1.5-Flash-002 demonstrates superior performance over Gemini-1.0-Pro in five out of six benchmark tasks. While Gemini-2.0-Flash-Exp consistently loses to Gemini-1.5-Pro-002, the margin is often narrow. Gemini-2.0-Flash-Exp outperforms Gemini-1.5-Flash-002 in 4 out of 6 tasks, primarily underperforming in tasks where progress must often remain stalled (Key-Chest, where no key might be found for many steps, and Windy Grid World, where after getting somewhat nearer to the target it becomes a battle to avoid being blown further away again). Finally, Gemini-2.0-Flash-Thinking-Exp-01-21 is improving upon that by also handling windy gridworld well while being the only model performing truly well on the temperature control task. However, also the thinking version of 2.0-Flash is underperforming on the Key-Chest exploration-oriented task. The task stands out as providing no signal for progress or change until the key is found, while maze has a smell of chess indicating nearness and the windy grid world features the distance to target. While the tasks were defined before the arrival of Gemini-1.5, and used to empirically nail down the agent workflow, they are now nearer to saturation with the later models where both 2.0-Flash-Exp-Thinking and Gemini-1.5-pro largely solves 5 out of 6.

\subsection{Social Simulation}
We implemented a multi-agent social simulation involving three characters (Alice, Bob, and Charlie) engaged in a game of catch, with an additional world entity introducing environmental events. Each character is controlled in its own stream, determining the character's actions and dialogue while maintaining consistency with the game state but without directly controlling other characters' behaviors. The world entity introduces contextual elements such as ice cream trucks and squirrels, enriching the environment without directly interfering with the core game mechanics. The characters observes a summary of the total story from their perspective, produced from the stream seeing all contributions, while the characters contribution is produced only seeing the own stream with the characters observation (worldview) contribution, action and update to the ball state.

The simulation runs for 25 time steps, with character responses evaluated against the current game state. Characters can perform actions such as throwing or catching the ball, accompanied by natural language dialogue. Each character starts with a consistency score of 5. When a character's contribution is inconsistent with the game state (e.g., attempting to throw a ball they don't possess), they receive one opportunity to revise their response. If the revised response remains inconsistent, a penalty is applied by decrementing their consistency score by one.

\begin{figure}[t]
   \centering
   \includegraphics[width=\columnwidth]{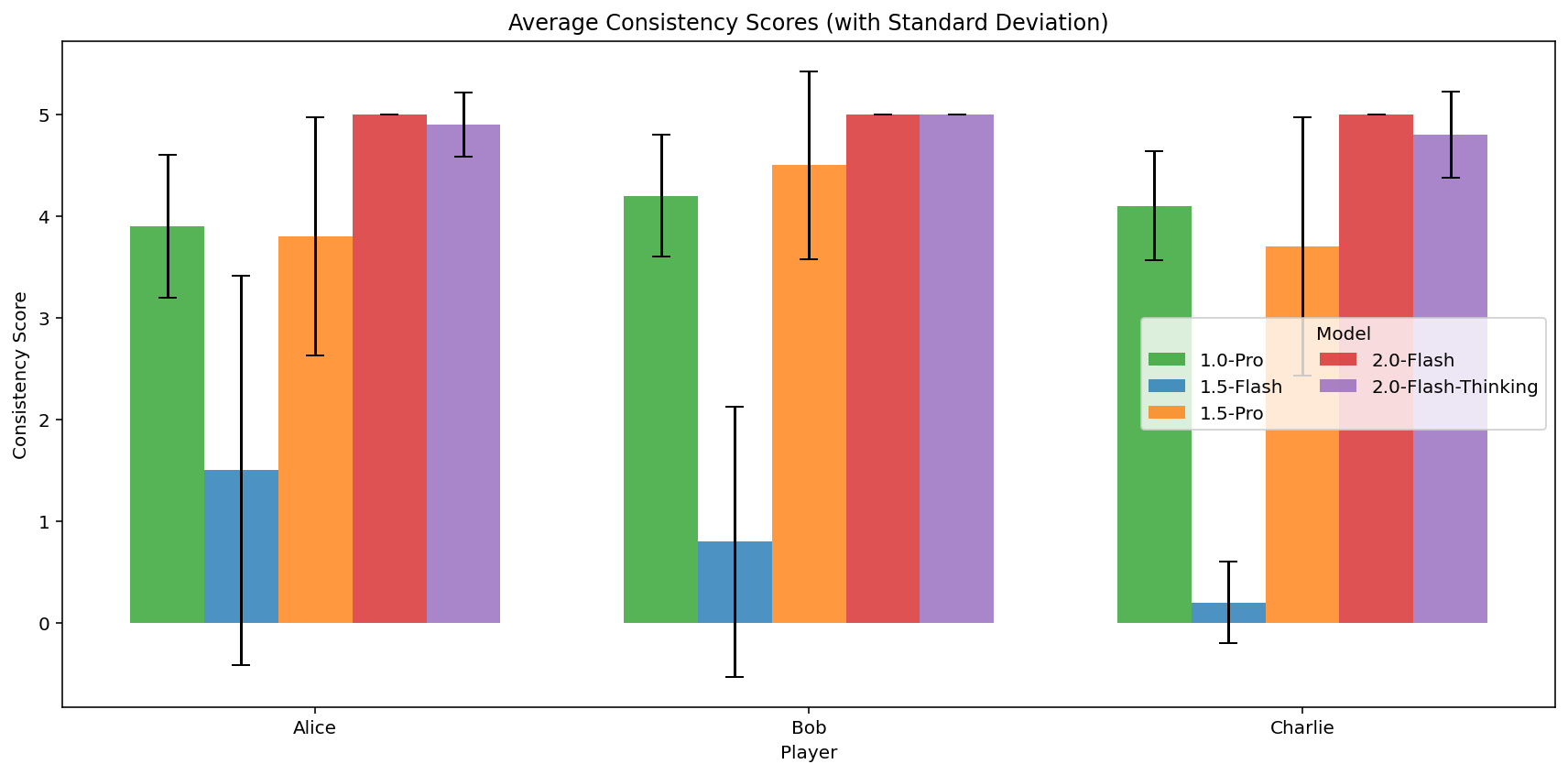}
   \caption{Inconsistency tracking across 10 runs of the social simulation. Each line represents a character's cumulative inconsistency count over 25 time steps, with variations between different model versions shown.}
   \label{fig:social_sim}
\end{figure}

We conducted 10 independent runs with the same range of models as above to assess their ability to produce a consistent simulation stream within the simulation streams framework. Figure~\ref{fig:social_sim} presents the consistency for each character across these runs, tracking when their responses required correction or incurred penalties. 

Gemini-2.0-Flash-Exp maintains complete consistency across each character's 250 decisions, and Gemini-2.0-Flash-Thinking-Exp makes only a handful of mistakes. Gemini-1.5-Pro-002 and Gemini-1.0-Pro show some occasional inconsistency, while Gemini-1.5-Flash-002 exhibits substantially more. Across all simulations, Gemini-2.0-Flash-Exp also demonstrates perfect adherence to generating exactly one additional simulation row in the correct format, resulting in substantially greater speed and economy due to fewer excess tokens and reduced need for resampling, which the framework requires when the generated response doesn't conform to the format.

\subsection{Market Simulation}

\begin{figure*}[]
    \centering
    \begin{subfigure}[b]{0.45\textwidth}
        \includegraphics[width=\columnwidth]{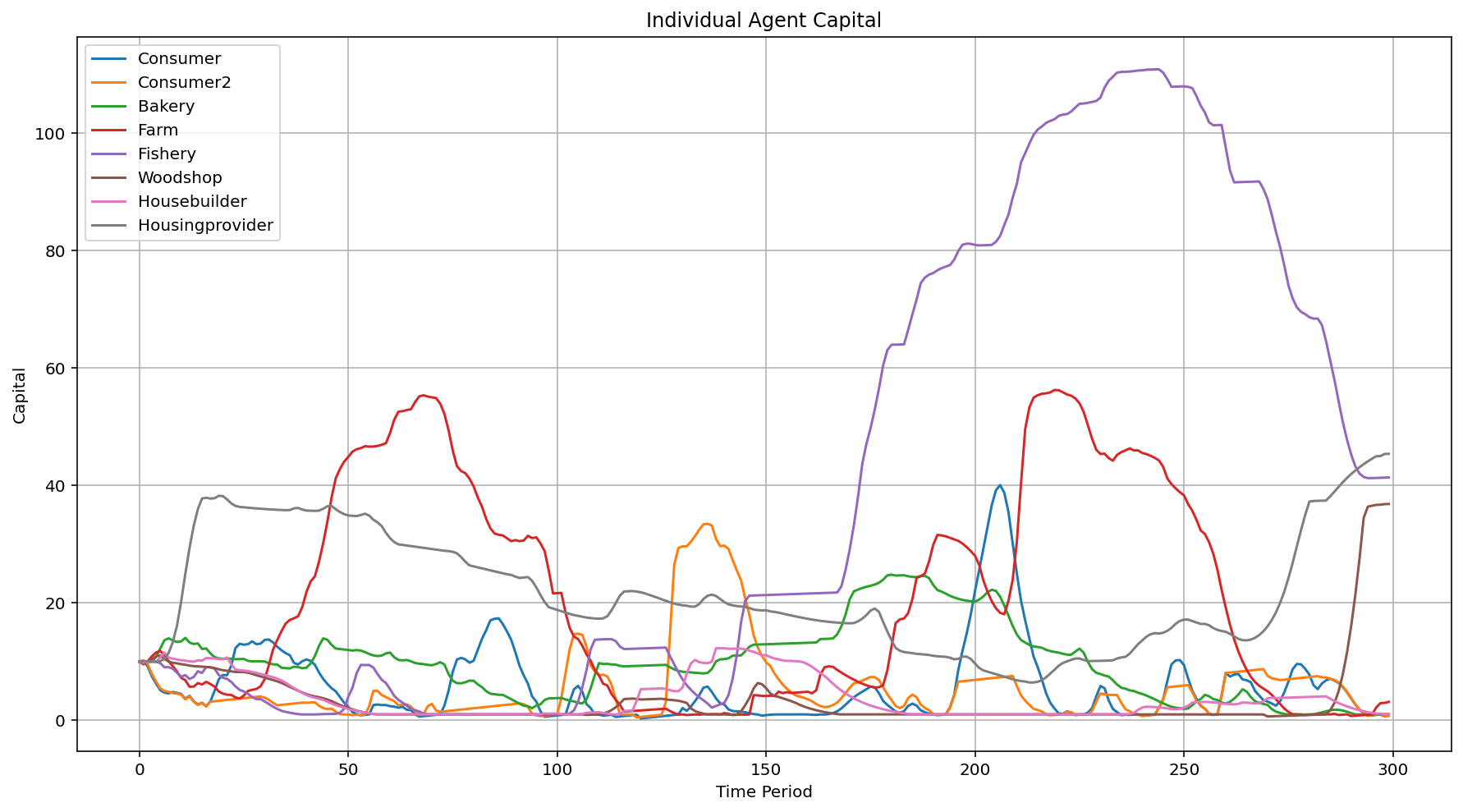}
        \caption{Capital distribution across market participants}
        \label{fig:capital}
    \end{subfigure}
    \hfill
    \begin{subfigure}[b]{0.45\textwidth}
        \includegraphics[width=\columnwidth]{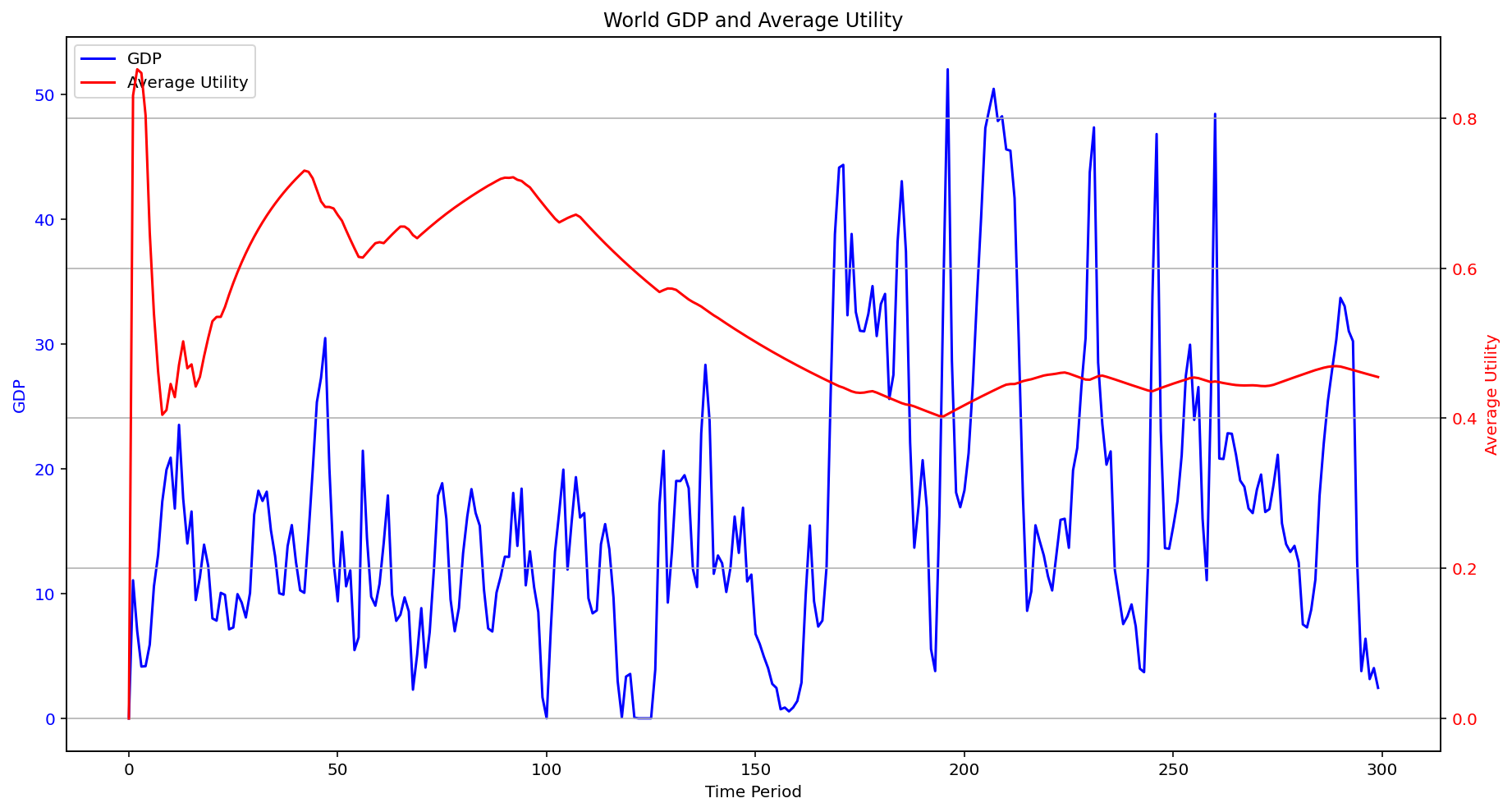}
        \caption{World GDP and average utility (leisure-based)}
        \label{fig:utility}
    \end{subfigure}
     \begin{subfigure}[b]{0.45\textwidth}
        \includegraphics[width=\textwidth]{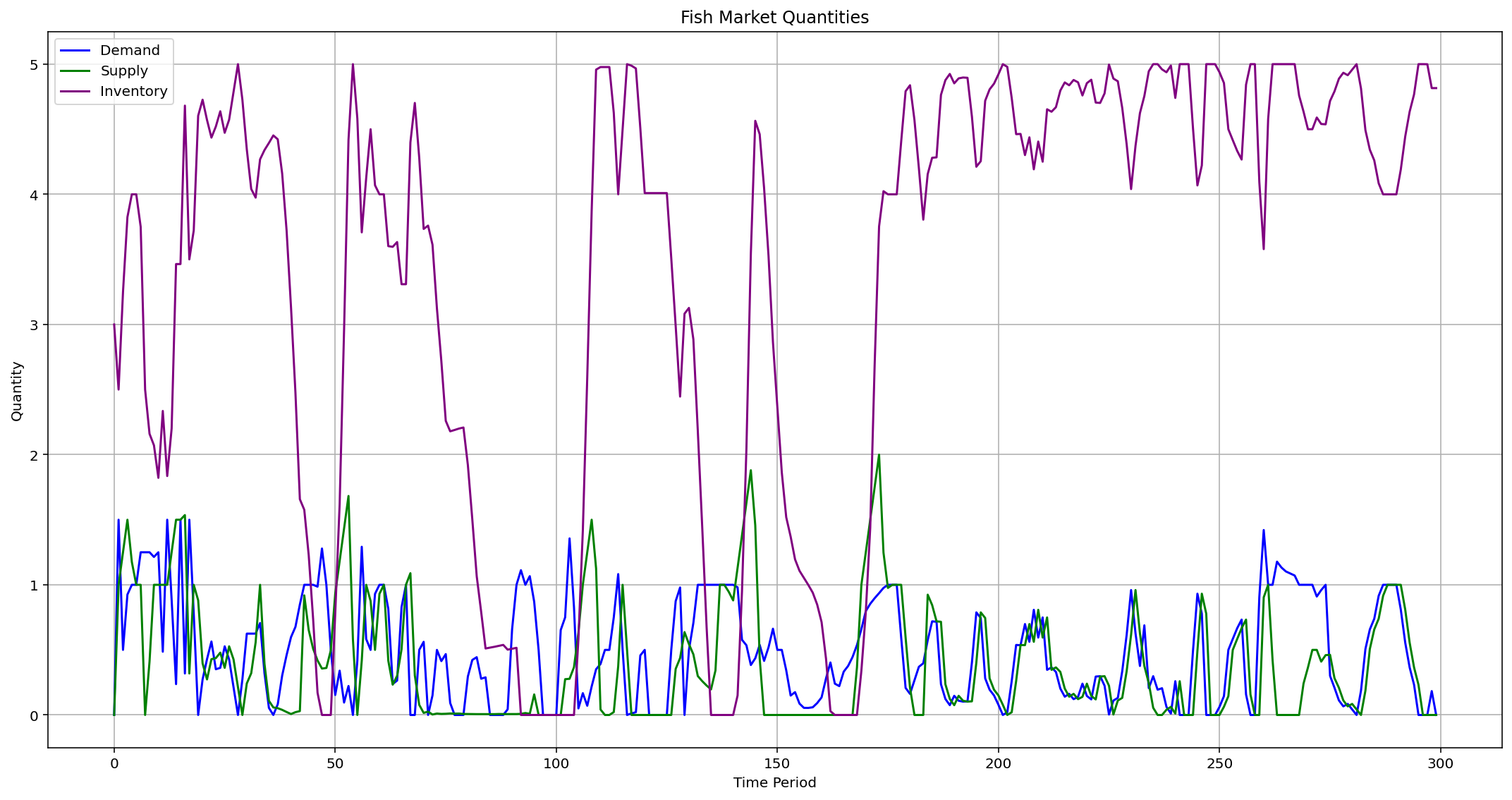}
        \caption{Fish market dynamics}
        \label{fig:fish}
    \end{subfigure}%
    \hfill
    \begin{subfigure}[b]{0.45\textwidth}
        \includegraphics[width=\textwidth]{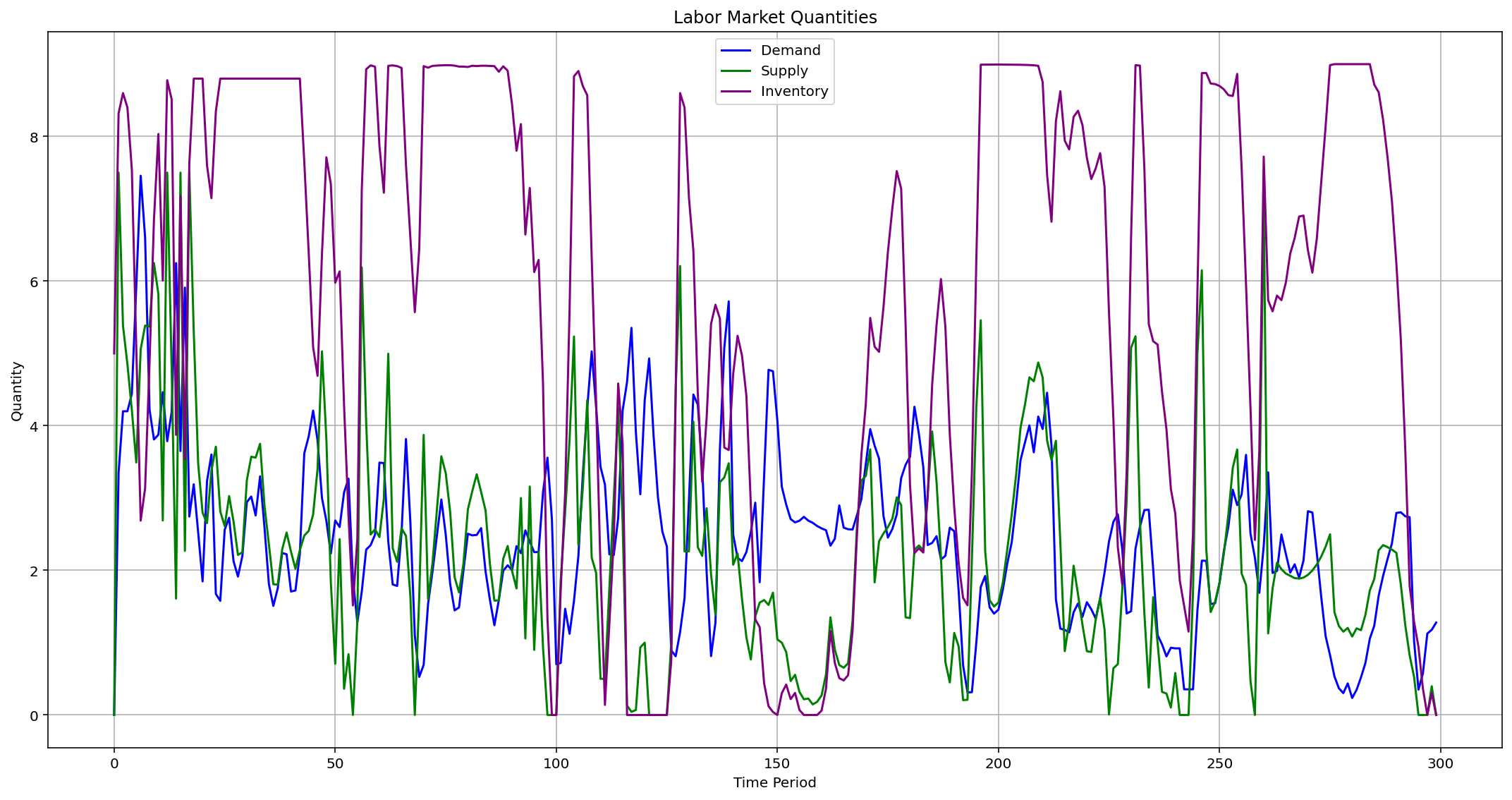}
        \caption{Labor market dynamics}
        \label{fig:labor}
    \end{subfigure}%
     
    \begin{subfigure}[b]{0.45\textwidth}
        \includegraphics[width=\textwidth]{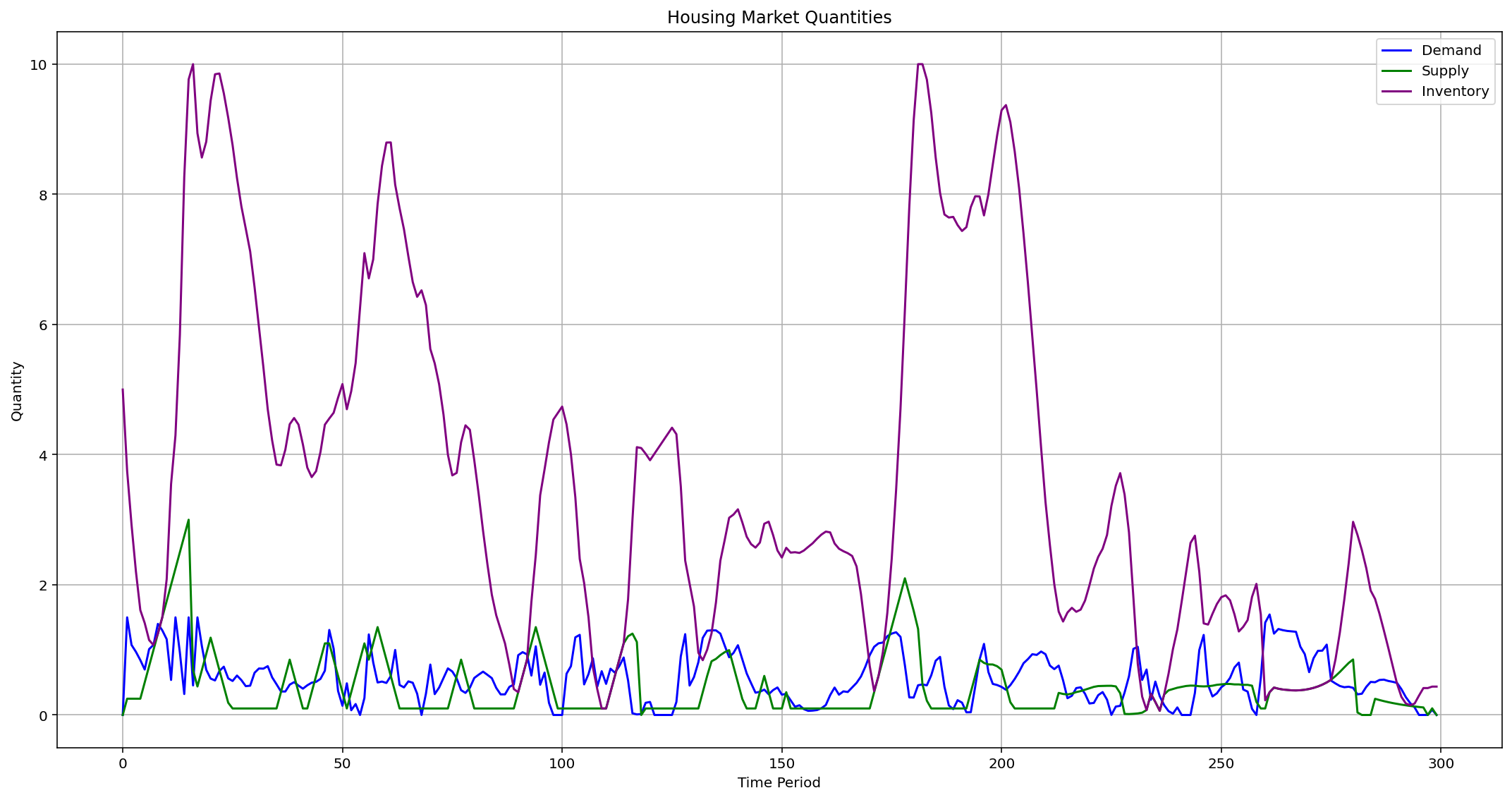}
        \caption{Housing market dynamics}
        \label{fig:housing}
    \end{subfigure}%
    \hfill
    \begin{subfigure}[b]{0.45\textwidth}
        \includegraphics[width=\textwidth]{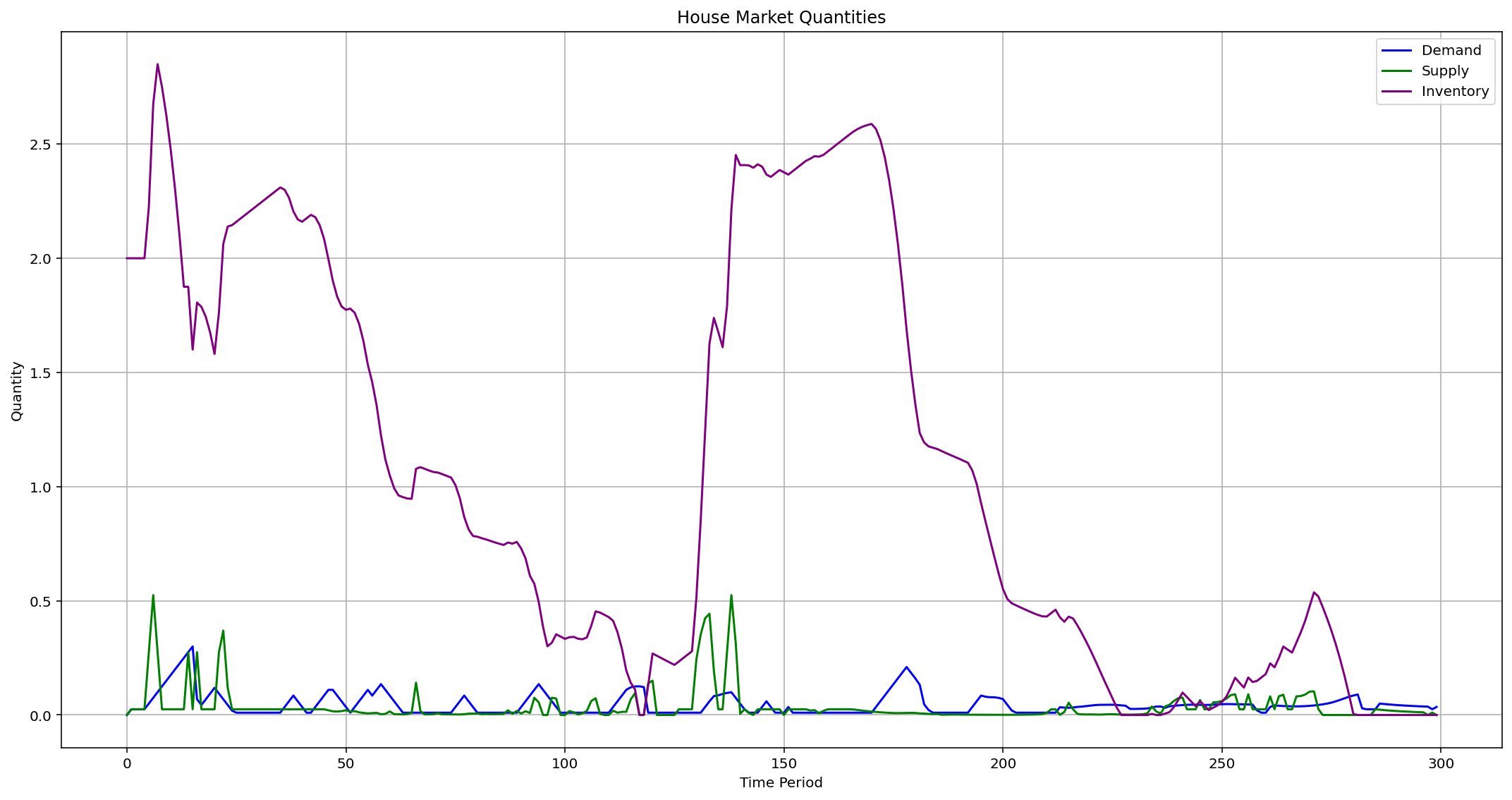}
        \caption{House building market dynamics}
        \label{fig:houses}
    \end{subfigure}

    \begin{subfigure}[b]{0.45\textwidth}
        \includegraphics[width=\textwidth]{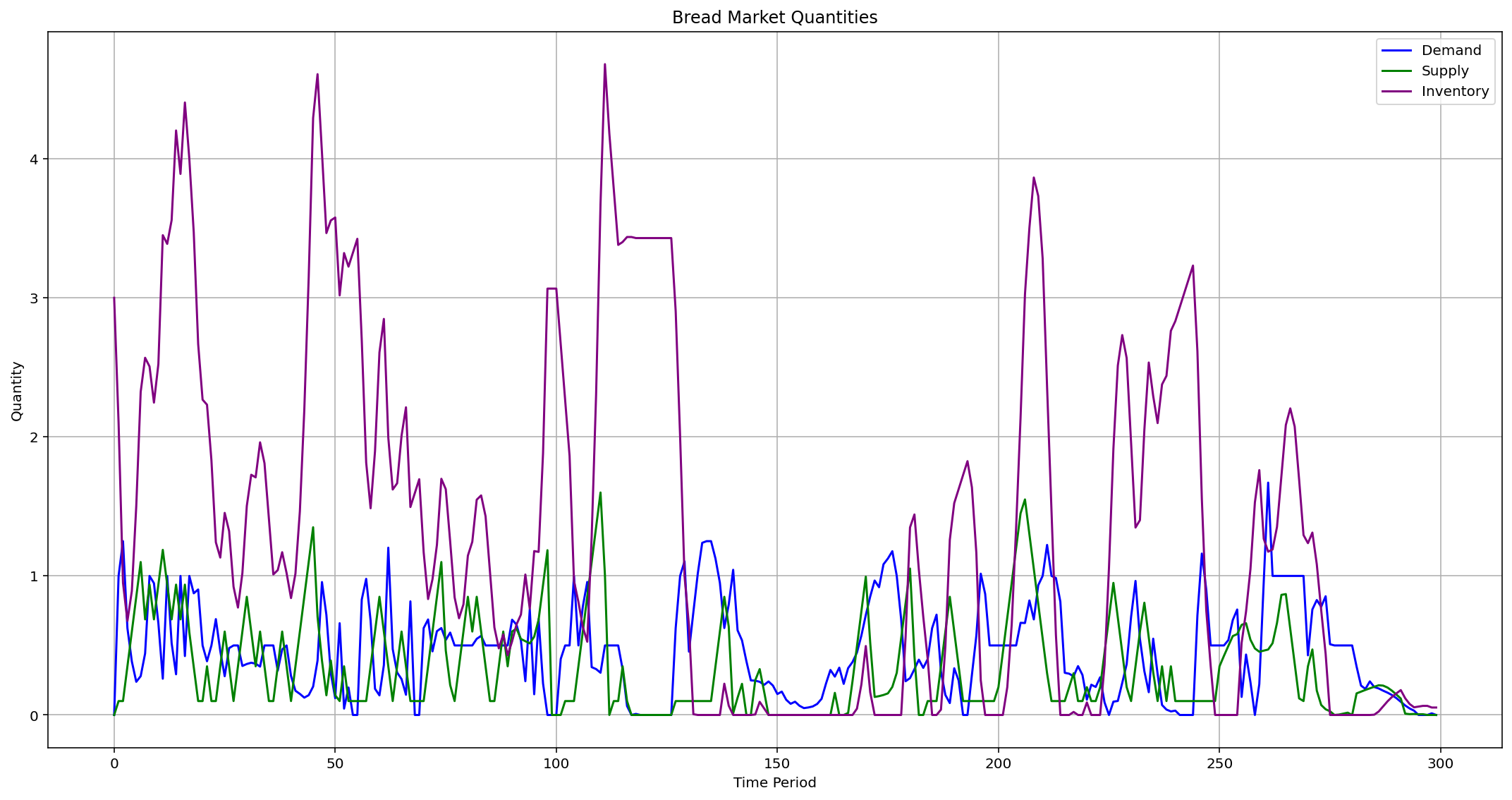}
        \caption{Bread market dynamics}
        \label{fig:bread}
    \end{subfigure}%
    \hfill
    \begin{subfigure}[b]{0.45\textwidth}
        \includegraphics[width=\textwidth]{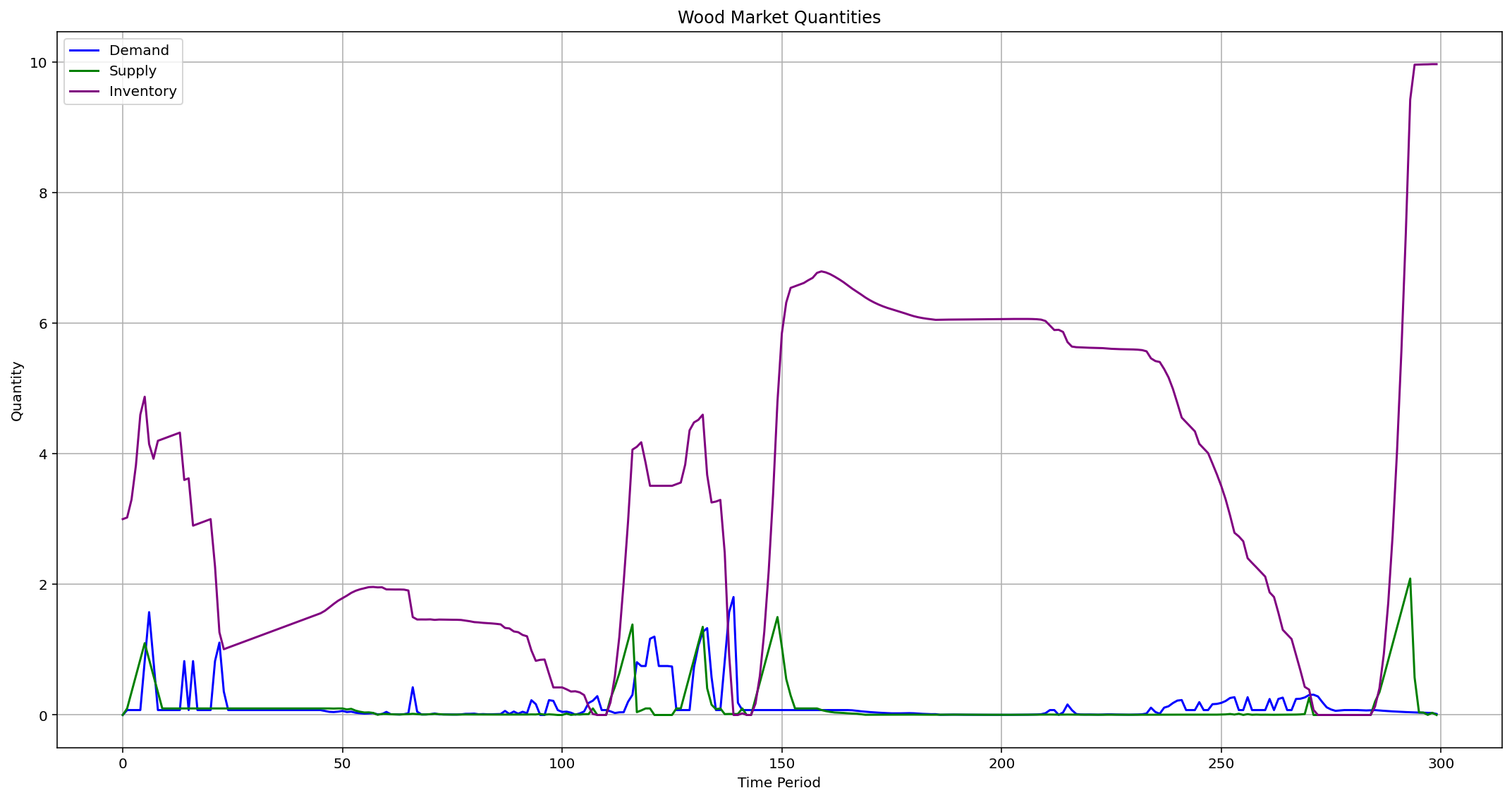}
        \caption{Wood market dynamics}
        \label{fig:wood}
    \end{subfigure}

    \begin{subfigure}[b]{0.45\textwidth}
        \includegraphics[width=\textwidth]{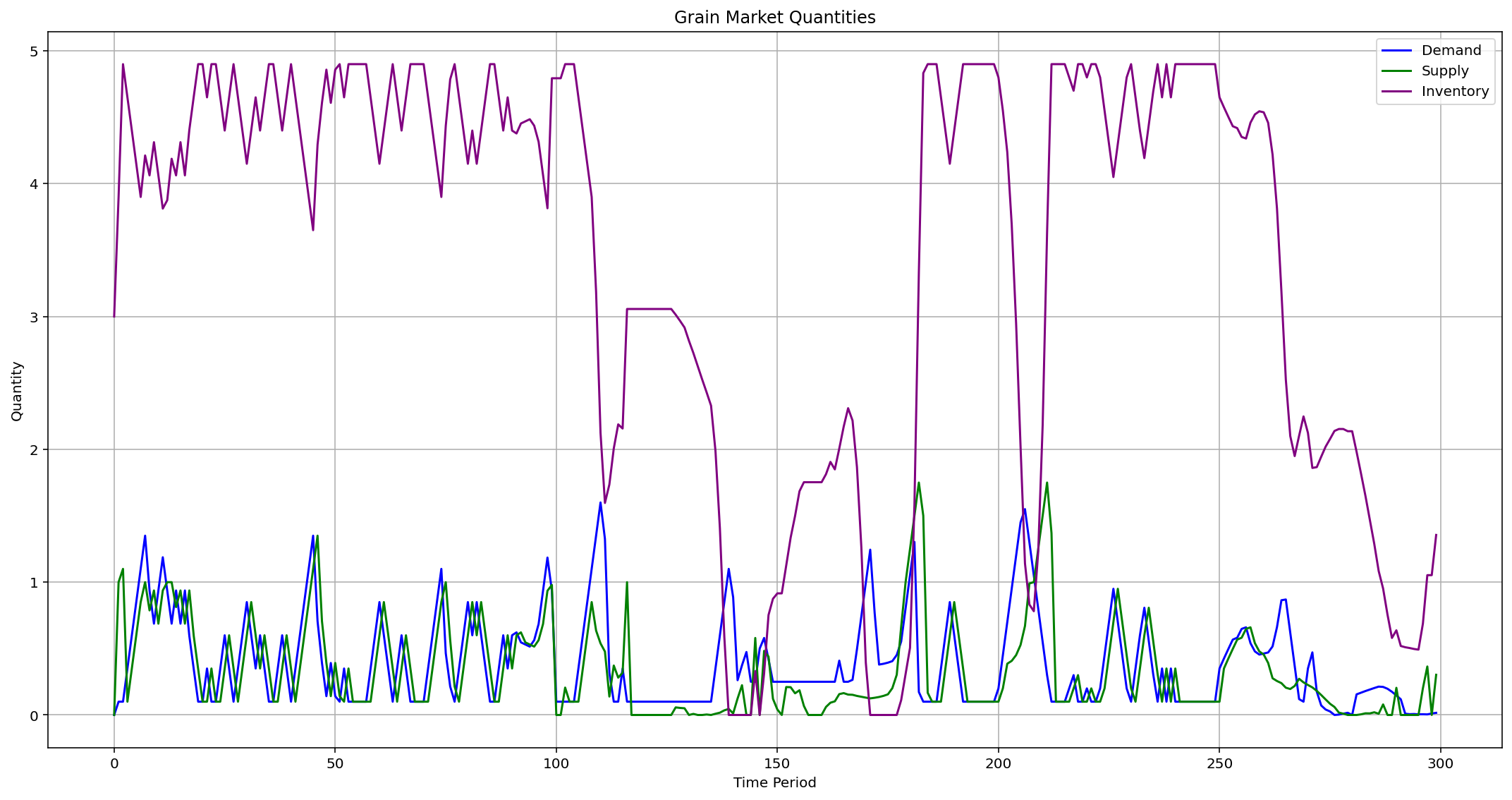}
        \caption{Grain market dynamics}
        \label{fig:grain}
    \end{subfigure}%
    \hfill
    \begin{subfigure}[b]{0.45\textwidth}
        \includegraphics[width=\textwidth]{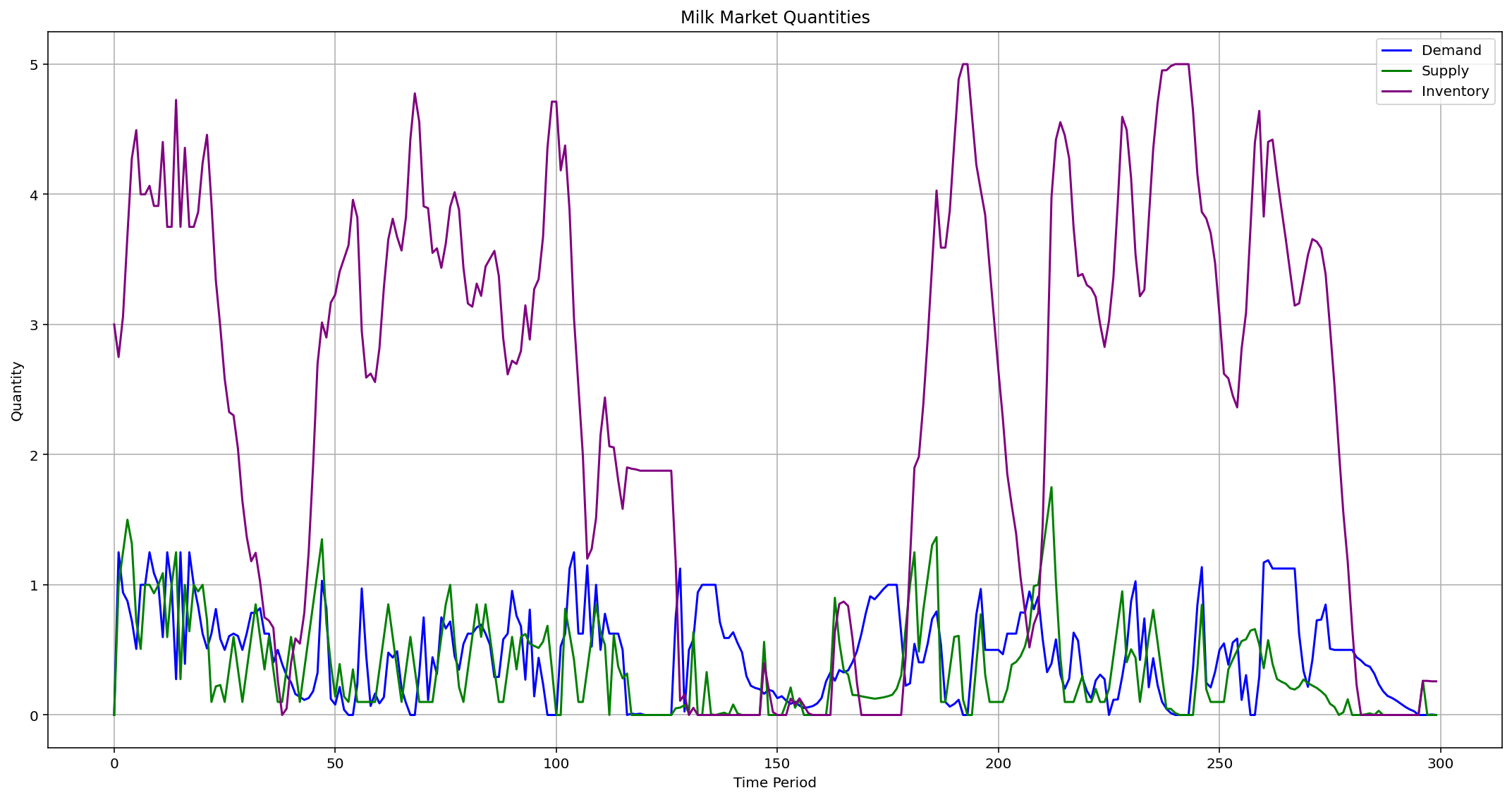}
        \caption{Milk market dynamics}
        \label{fig:milk}
    \end{subfigure}
    \caption{Market dynamics for various goods and resources over 250 time steps. Capital distribution, GDP as well as supply, demand and inventory levels for some goods.}
    \label{fig:market_dynamics}
\end{figure*}

Our simulated economy consists of eight key markets: housing, houses, bread, wood, grain, milk, fish, and labor. The producing entities and the consumers each have their own experience stream, but where the current value of relevant variables such as price and inventories is being displayed. There is also a stream for the market maker setting prices, observing the demand and inventory variables for a good before setting the price. The key variables are affected by the decisions of different entities but they only observe this by displaying the variables through a formula, i.e. the relevant parts of the current state.

Figure \ref{fig:market_dynamics} provides metrics of overall economic development including total GDP and average utility, where utility is calculated as 0.2*(10-labor inventory) when labor inventory exceeds 5 (representing available leisure time) and 0 otherwise. It also presents the dynamics of different goods, with plots of supply, demand and inventory levels. We plotted the simulation results for 250 time steps, observing oscillating market dynamics and adaptive behaviors.

\section{How to write a Simulation Stream program}

Writing a Simulation Stream program involves defining an initial state and a list of operators. While the formal definition was presented in Section 2, here we demonstrate the concrete implementation. We first demonstrate the straight forward approach that directly writes the list of operators, and then how we can use the ECS approach to make that more scalable. 

\subsection{State Initialization}

The program begins with initialization of the state variables:

\begin{verbatim}
initial_state = {
    'time': 0,
    'location_x': 0,
    'location_y': 0,
    'move_x': 0,
    'move_y': 0,
    'previous': [],
    'cheese_found': False,
    'planning': False,
    'movement': False
}
\end{verbatim}

\subsection{Defining Operators}

The core of a Simulation Stream program is a list of operators. Each operator specifies how variables change and when to involve the language model. Here is an example operator list showing part of the cheese-finding program referenced earlier:

\begin{verbatim}
operators = [
    {
        'id': 'Time',
        'formula': 'time = time + 1',
        'use_lm': False,
        'query': {},
        'planning': True,
        'movement': True,
        'summary': True,
        'next': 'Objective'
    },
    ...
    {
        'id': 'Summary',
        'formula': 'summary = ...',
        'use_lm': 'time > 1',
        'query': {'summary': True},
        'planning': True,
        'movement': False,
        'summary': True,
        'next': 'Time'
    }
]
\end{verbatim}
where ... in the summary operator should be a string that shows a useful summary of the first step in that simulation, a step where everything has been determined by initialization and not yet including the large language model that starts generating choices in the second block.

Each operator in the list contains several key fields:
\begin{itemize}
    \item \texttt{id}: A unique identifier for the operator
    \item \texttt{formula}: The formula that assigns a value to a variable
    \item \texttt{use\_lm}: A boolean or condition specifying when to use the language model
    \item \texttt{query}: A dictionary specifying which flags to match when selecting context
    \item \texttt{planning}, \texttt{movement}: Boolean flags indicating the operator's categories
    \item \texttt{next}: The identifier of the operator to execute next
\end{itemize}

When executed, these operators generate the output stream shown in Figure 2, where some rows are language model-generated while others come from deterministic formulas. The flags (\texttt{planning}, \texttt{movement}) are used to mark rows in the output stream, enabling subsequent operators to selectively query relevant context through their \texttt{query} dictionaries.

For example, when the \texttt{Summary} operator executes, it will only receive context from previous rows where \texttt{summary=True}, ensuring it has access to relevant information without being overwhelmed by unrelated details.

The program's execution follows the sequence defined by the \texttt{next} fields, with each operator either executing its formula directly or invoking the language model based on its \texttt{use\_lm} condition. This produces a structured output stream that maintains format consistency while leveraging language model capabilities where appropriate.

\subsection{Entity-Component Organization}

The same program can be organized using the Entity-Component-System (ECS) approach, which provides better scalability and reuse. For our cheese-finding example:

\begin{verbatim}
entities = {
    'world': [
        'heading'
    ],
    'agent': [
        'planning',
        'movement',
        'summary'
    ]
}

variables = {
    'heading': {
        'time': 0,
        'objective': None
    },
    'movement': {
        'location_x': 0,
        'location_y': 0,
        'move_x': 0,
        'move_y': 0
    },
    'planning': {
        'high_level_plan': None,
        'movement_plan': None
    },
    'summary': {
        'previous': [],
        'cheese_found': False,
        'first_summary': "..."
    }
}
\end{verbatim}

This organization splits our previous operator list into components, each associated with specific entities:

1. The world entity only has a simple heading component
2. The agent entity manages planning, movement, and summary generation

Each component's operators maintain their original functionality while being organized in a more modular way. This approach makes it easier to add new entities (like additional agents) or components (like communication between agents) to the simulation.

\section{Implementation Details}

The implementation\footnote{\href{github.com/google-deepmind/simulation_streams}{\url{github.com/google-deepmind/simulation_streams}}} consists of a modular Python codebase that implements an Entity-Component-System (ECS) architecture for simulation management. 

\subsection{Core Architecture}

The codebase is structured around a Flask-based web application that provides an interactive web interface for simulation management as well as a command line option. The server handles HTTP requests that drive UI updates and simulation state changes through JavaScript code injection. The main components are:

\noindent A Flask application server (\texttt{app.py})
\\An ECS editor backend (\texttt{editor.py})
\\Simulation utilities (\texttt{simulation\string_utils.py})
\\An expression evaluator (\texttt{expressions.py})
\\A web interface (\texttt{templates/index.html})
\\A configs directory with simulation configuration files
\\A venv setup file (\texttt{setup.sh})

\subsection{Key Components}

\subsubsection{Flask Application Server}
The application server handles:
\begin{itemize}
    \item Entity-Component management endpoints
    \item Simulation control and execution
    \item Continuous metric tracking and visualization
\end{itemize}

The server uses route decorators to define endpoints that interface with the ECS editor:

\subsubsection{ECS Editor}
The \texttt{ECSEditor} class serves as the core backend component, managing:
\begin{itemize}
    \item Entity-Component definitions
    \item Variable and operator management
    \item Simulation state
    \item GUI state synchronization
\end{itemize}

The editor keeps the simulation definition as a  structured dictionary:

\begin{verbatim}
self.ecs = {
    'entities': {},
    'variables': {},
    'systems_definitions': {}
}
\end{verbatim}

\subsubsection{Expression Evaluator}
The expressions module provides a secure evaluator for executing simulation expressions, incorporating:
\begin{itemize}
    \item Mathematical functions from \texttt{math}
    \item Built-in Python functions
    \item String manipulation methods
    \item Statistics functions
    \item Custom extensions
\end{itemize}

The evaluator uses the \texttt{SimpleEval} library.

\subsubsection{Simulation Utilities}
The utilities module provides core simulation functionality including for:
\begin{itemize}
    \item The simulation stream generator
    \item State transitions and output stream generation
    \item Querying history for sub-stream generation
\end{itemize}

\subsection{Command line options}

The system supports various configuration options through command-line arguments:

\begin{description}[labelwidth=\widthof{\textbf{--output-file}}, leftmargin=!]
\item[\textbf{\texttt{ecs\string_file}}] (optional) \
Path to the ECS configuration file
\item[\textbf{\texttt{--web}}]
Flag to launch the web interface \\
\emph{Default: False}
\item[\textbf{\texttt{--steps}}]
Number of simulation steps \\
\emph{Default: 10}
\item[\textbf{\texttt{--metrics}}]
Metrics tracking file path \\
\emph{Default: None}
\item[\textbf{\texttt{--index}}]
Variable initialization index \\
\emph{Default: 0}
\item[\textbf{\texttt{--model}}]
LLM model identifier \\
\emph{Default: gemini-1.5-pro}
\item[\textbf{\texttt{--api-key}}]
Model API access key \\
\emph{Default: Empty string}
\item[\textbf{\texttt{--output-file}}]
Query results output path \\
\emph{Default: None}
\end{description}

\bibliography{main}

\end{document}